\documentclass{article}

\usepackage[final]{neurips_2021}

\usepackage[utf8]{inputenc} 
\usepackage[T1]{fontenc}    
\usepackage{hyperref}       
\usepackage{url}            
\usepackage{booktabs}       
\usepackage{amsfonts}       
\usepackage{nicefrac}       
\usepackage{microtype}      
\usepackage{xcolor}         

\setcitestyle{numbers,comma,square}

\title{Can machines learn to see without visual databases?}

\author{%
    Alessandro Betti$^{1}$, Marco Gori$^{1,2}$, Stefano Melacci$^{1}$, Marcello Pelillo$^{3}$, Fabio Roli$^{4}$ \\
  $^{1}$DIISM, University of Siena, Siena, Italy \\
  $^{2}$Maasai, Universit\`{e} C\^{o}te d'Azur, Nice, France \\
  $^{3}$DAIS, Ca' Foscari University of Venice, Venice, Italy \\
  $^{4}$DIEE, University of Cagliari, Cagliari, Italy \\
  \texttt{alessandro.betti2@unisi.it, \{marco,mela\}@diism.unisi.it,} \\ \texttt{pelillo@unive.it, roli@unica.it} \\
}

\begin{document}

\maketitle

\begin{abstract}
This paper sustains the position that 
the time has come for thinking of learning machines that
conquer visual skills in a truly human-like context, where a few human-like object supervisions are given by vocal interactions and pointing aids only. This likely requires new foundations on computational processes of vision with the final purpose of involving machines in tasks of visual description by living in their own visual environment under simple man-machine linguistic interactions.  The challenge consists of developing machines that learn to see without needing to handle visual databases. This might open the doors to a truly orthogonal competitive track concerning deep learning technologies for vision which does not rely on the accumulation of huge visual databases.
\end{abstract}

\section{Introduction}
\label{sec:intro}
The outstanding results of deep learning of the last decade have paved the road to the creation of powerful neural network-based applications in different domains \cite{alexnet,alphago,gpt3}. The performance of deep networks inherently depends on the quality of the datasets that are used during the training stage, usually large scale \cite{imagenet,Tan_2020_CVPR}.
As a matter of fact, data collection and data management have become crucial processes for which companies need to plan investments, if they aim at improving different parts of their production chains using deep learning approaches. 
Collecting data also involves costly annotation procedures to generate the ground truths that the machine exploits to learn the properties of a novel task, or that are also used to measure the performance of the machine itself. The quality of the annotations must be ensured by specific checking activities that cannot be easily fully automatized.
In order to cope with the costs of massively supervising large databases, modern solutions are strongly based on neural networks that are first pretrained on large public data collections, exploiting transfer learning and well-tuned adaptations to comply with the target task \cite{oquab2014learning}. The recent emphasis on self-supervised learning has created the conditions for developing robust representations without human intervention, even if there is still the need of huge databases \cite{DBLP:journals/corr/abs-2104-14294}. In a nutshell, ``data'' remain a crucial aspect that requires further attention and studies to go beyond or to partially overcome the aforementioned issues and others. Ensuring data quality is becoming the fuel of AI-based software systems.

We focus on the case of machines that are designed to conquer basic visual skills, learning representations that, afterwards, can be used as the starting block to learn predictors oriented toward more specific domains. Using data from other tasks to gain these basic skills, as in the case of image classification datasets, or huge collections of unlabeled images taken from the web, or mixing heterogeneous datasets, is what currently plays a crucial role in machine learning. In some cases, those collections are scarcely interesting in themselves (random images from the web). In other cases, they might involve critical privacy issues whenever they store a few pictures or videos taken from contexts that were not meant to be shared or memorized. We refer to these data as {\em Perceptual Data Collections}, and they could be about different perceptual information, such as in in the case of vision and speech. Differently, there exists data collections  storing information that can actually boost the performance of machine learning apps, introducing knowledge that goes beyond what might be important to gain basic perceptual skills, such as higher level of abstraction, or specializing the machine in a specific task.  These databases have value independently of whether they are used or not in machine learning, and they store precious information that is of crucial importance per se. Hence, it is mandatory to focus on their maintenance according to professional data engineering schemes that also contribute significantly to improve the quality of machine learning models.


This paper sustains the position that the time has come 
to carefully address the above mentioned different 
roles of data collection. In particular, we consider the case of  
{\em Perceptual Data Collections} for computer vision and 
stimulate a discussion on their actual need for developing
machines capable of exhibiting the current visual skills. 
The spectacular recent results of deep learning are in fact
delaying a fundamental question that, soon or later will likely
addressed:
\begin{quote}
{\em Can machines learn to see without visual databases like
ImageNet? Can we devise solutions for acquiring visual skills from the simple on-line processing of video signal and ordinary human-like scene descriptions?}
\end{quote}
Couldn't it be the case that we have been working on a problem that is---from a computational point of view---remarkably different and likely more difficult with respect to the one offered by Nature? Sensory information is intrinsically characterized by a natural temporal development of the data, that carries precious information. The case of vision (but also of speech, for example) is an idiomatic example that concretely tells that we learn from data that follows a precise temporal order, and the learning process naturally proceeds over time. In other words, we do not learn from a huge dataset of shuffled images. Supervision does not consist of a detailed labeling of each single piece of information that is acquired, since the way we learn is not only due to a passive exposition to (eventually) pre-buffered data. In nature, animals gain visual skills without storing their visual life!
Why can they afford to see without accessing a previously stored visual database? Is there a specific biological reason that cannot be captured in machines? This paper sustains the position that those skills can be likely gained also by machines and proposes the challenge of learning to see without using visual collections. This argument is supported by the belief that also in nature visual skills obey information-based principles that biology simply exploits to achieve the objectives required by living in a certain environment. We suggest opening a challenge where machines live in their own visual environment and  adopt a natural learning protocol based on the continuous on-line process of the acquired video signal and on ordinary humans interaction for describing visual scenes.

\section{Learning in visual environments}
We can think of shifting from the idea of labeling pre-buffered data to actively interacting with the artificial agent that is being trained, by exposing it to a form of supervision that depends on the time in which it is provided. In this context, the agent can take the initiative of asking for a supervision and also the (human) supervisor can take the initiative of interacting. This affects dramatically  the current labelling procedures, since the interaction takes place in a truly human-line context, thus providing labels that are more appropriate at a specific time instant.
As we involve the crucial role of time, this way of framing the learning process leads to some very stimulating challenges. For example, 
one soon realizes that there is huge difference between the 
capability of detecting a specific object and recognizing the 
abstract category that is attributed to objects. 
Humans can recognize objects identity regardless of the different poses in which they come, also in presence of environmental noise. We conjecture
that such a skill is inherently connected with the capability of 
learning visual features that are invariant under motion. 
Such an invariance is in fact centered around the object itself and, as such, it does reveal  its own features in all possible poses gained during motion. 
However, the development of the learning capabilities to gain related 
visual skills can only aspire to  discover objects' identity.
Humans, and likely most animals, also conquer a truly different understanding of visual scenes that goes beyond the conceptualization with single object identities that involve their shape and appearance. In early Sixties, James J. Gibson coined the notion of {\em affordance} in~\cite{Gibson1966}, even though 
a more refined analysis came later in~\cite{plasticq}. 
A chair, for example, has the affordance of seating a human being, but can have  many other potential uses. We can climb on it to reach a certain object or we can use it to defend ourselves from someone armed with a knife.
It was also early clear that no all objects are characterized
by the affordance in the same way.  In~\cite{plasticq}, the distinction between {\em attached} and {detached objects}
clearly indicates that  some can neither be moved nor easily broken. While our home can be given an affordance, it is attached to the earth and cannot be grasped. Interestingly, while a small cube is  graspable, huge cubes are not. However, a huge cube can still be graspable by an appropriate ``handle''.  Basically,  one can think of object perception in terms of its shape and appearance.  However, this is strictly connected to the object identity, whereas most common and useful 
ways of  gaining the abstract notion of objects likely comes from their affordance. Hence, gaining abstraction is significantly different from conquering the skills of detecting the identity of a specific object.

We summarize the main elements of the previous discussion into the following problems.
\begin{itemize}
    \item \textit{Object recognition in visual environments from human vocal interactions.}
Design and implementation of visual agents that can act in natural visual environments for both surveillance and ego-centric vision tasks. The agents are also expected to return pixel-level maps to segment the objects. The challenge consists of gaining object recognition skills under ordinary  vocal human interactions where the object supervision is provided.

\item \textit{Object identity vs. abstract categorization.} 
Design and implement visual agents that are capable of distinguishing the notion of a specific object with respect to its abstract category. For example,  the agent is expected to recognize “my own chair” with respect to the general abstract notion of chair.

\item \textit{Feature transmission.}
Based on the intuition behind transfer learning,  capitalize from motion invariance features that are developed in different visual environments to devise representations for firing subsequent learning processes suited to face specific problems.
\end{itemize}

In this paper we propose the challenge of attacking the above problems without involving visual databases. This might led to a paradigm-shift on methods for learning to see, with important ethical impacts on data accumulation.

\section{Benchmarks and crowdsourcing}
So far, computer vision has  strongly benefited from the massive diffusion of
benchmarks which, by and large, are regarded as fundamental tools for performance evaluation \cite{imagenet}.
However, it is clear that they are very well-suited to support the statistical machine learning
approach based on huge collections of labeled images. However, in this paper we propose to explore a truly different framework for performance evaluation. 
The emphasis on video instead of images does not necessarily lead to the 
accumulation of huge collections of video. Just like humans, machines are expected 
to ``live in their own visual environment''' and can be evaluated on-line.
This needs an in-depth re-thinking of nowadays benchmarks.
They bears some resemblance to the influential testing movement in psychology which has its roots in 
the turn-of-the-century work of Alfred Binet on IQ tests~(\cite{Binet1916}).
Both cases consist in attempts to provide a rigorous way of assessing the performance or the aptitude of a 
(biological or artificial) system, by agreeing on a set of standardized tests which, 
from that moment onward, become the ultimate criterion for validity. 
On the other hand, it is clear that the skills of any visual agent can be quickly evaluated 
and promptly judged by humans, simply by observing its behavior. 
How much does it take to realize that we are in front of person with visual deficits? 
Do we really need to accumulate tons of supervised images for assessing the quality of
a visual agent? ImageNet~\cite{imagenet} is based on crowdsourcing. 
Couldn't we also use crowdsourcing as  a {\em performance evaluation scheme}?
People who evaluate the performance could be properly registered so as to limit 
spam. 
Scientists in computer vision could start following a sort of 
{\em en plein air movement}.\footnote{Here, the term is used  to  mimic 
the French Impressionist painters of the 19th-century and, more 
generally, the act of painting outdoors} This term suggests that visual agents should be evaluated 
by allowing people to see them in action, virtually opening the doors of research labs. 

The complexity of the problem at hand suggests that shifting computer vision challenges
into the wild deserves attention. We can benefit dramatically from on-going virtual visual 
environments for supporting the experiments. In particular, we have already been experimenting
the {\tt SAILenv} platform for our preliminary experiments~\cite{meloni2021sailenv}.
We have also in the process of gaining experience with {\tt ThreeDWorld}
\cite{DBLP:journals/corr/abs-2007-04954}. The remarkable progress of graphic technologies
make these and other platforms more and more photorealistic, since we can perform very accurate evaluations
and reproduce assessment schemes that might resemble some of nowadays benchmarks. 


	Needless to say, modern computer vision has been fueled by the availability of huge labeled  image collections, which clearly shows the fundamental role played by pioneering projects in this direction (see e.g.~\cite{imagenet}). However, many questions posed in this paper will likely be better addressed only when  scientists will put more emphasis on the en plein air environment. In the meantime, we remark our claim that the experimental setting needs to move to virtual visual environments. Their photorealistic level along with explosion of the generative capabilities make these environment just perfect for a truly new performance evaluation of computer vision. The  advocated crowdsourcing approach might really change the way we measure the progress of the discipline.

\section{Ethical issues}
The cost of attaching labels to the collected data remains a key challenge in modern machine learning-based solutions, and many applications rely on transfer learning, using a pre-trained model as the basis for further computations and to face other scarcely supervised learning tasks \cite{oquab2014learning}. On one hand, this speeds-up the development and it usually yield improved performances, on the other hand, pre-trained models inherit the biases induced by the data they were trained on, that might not be a desirable property. Another key issue with the labeling process is also the one of providing labels to those examples that are expected to carry the most important pieces of information for the learning task. In fact, measuring the labeling quality simply counting the number of supervised examples should not be the only indicator to use when analyzing the collected data. 
Learning from scratch in an interactive manner avoids the needs of depending on external models or data, and it not necessarily requires to store the information that is collected during the agent life. This is of crucial importance when considering privacy issues that are easily triggered when acquiring data in a private context. For example, think about an agent that processes the data collected from the camera/microphone of a private smartphone. Even when neglecting the privacy issues if the processing does not take place in the mobile device, there is still a huge privacy issue in the way data and interactions are stored.

Unlike the dominant trend of devising computer vision technologies thanks to the accumulation of huge visual databases, on the long run, learning without visual databases  might become an important requirement for the massive adoption of computer vision technologies. This might have a significant  ethical impact and could stimulate novel developments in top level computer vision labs.
\section{Conclusions}
\label{sec:concl}
This paper sustains the position that the time has come to think of machines that can learn to see without accessing visual collections. The new learning protocol assumes that machines live in their own environment and can access object supervisions upon request or from the (human) supervisor initiative. When following this new framework, we will likely end up into an intriguing social context in which learning agents will pose supervising questions to other agents which have already gained good skills under a truly on-line scheme.  

\begin{ack}
This work was partly supported by the PRIN 2017 project RexLearn, funded by the Italian Ministry of Education, University and Research (grant no. 2017TWNMH2).
\end{ack}

{\small
\bibliography{References,nn}

\begin{thebibliography}{10}

\bibitem{alexnet}
Alex Krizhevsky, Ilya Sutskever, and Geoffrey~E Hinton.
\newblock Imagenet classification with deep convolutional neural networks.
\newblock {\em Advances in neural information processing systems},
  25:1097--1105, 2012.

\bibitem{alphago}
David Silver, Aja Huang, Chris~J Maddison, Arthur Guez, Laurent Sifre, George
  Van Den~Driessche, Julian Schrittwieser, Ioannis Antonoglou, Veda
  Panneershelvam, Marc Lanctot, et~al.
\newblock Mastering the game of go with deep neural networks and tree search.
\newblock {\em Nature}, 529(7587):484--489, 2016.

\bibitem{gpt3}
Tom~B Brown, Benjamin Mann, Nick Ryder, Melanie Subbiah, Jared Kaplan, Prafulla
  Dhariwal, Arvind Neelakantan, Pranav Shyam, Girish Sastry, Amanda Askell,
  et~al.
\newblock Language models are few-shot learners.
\newblock In {\em Advances in Neural Information Processing Systems}, 2020.

\bibitem{imagenet}
Jia Deng, Wei Dong, Richard Socher, Li-Jia Li, Kai Li, and Li~Fei-Fei.
\newblock Imagenet: A large-scale hierarchical image database.
\newblock In {\em 2009 IEEE conference on computer vision and pattern
  recognition}, pages 248--255. Ieee, 2009.

\bibitem{Tan_2020_CVPR}
Mingxing Tan, Ruoming Pang, and Quoc~V. Le.
\newblock Efficientdet: Scalable and efficient object detection.
\newblock In {\em Proc. of the IEEE/CVF Conf. on Computer Vision and Pattern
  Recognition (CVPR)}, June 2020.

\bibitem{oquab2014learning}
Maxime Oquab, Leon Bottou, Ivan Laptev, and Josef Sivic.
\newblock Learning and transferring mid-level image representations using
  convolutional neural networks.
\newblock In {\em Proceedings of the IEEE conference on computer vision and
  pattern recognition}, pages 1717--1724, 2014.

\bibitem{DBLP:journals/corr/abs-2104-14294}
Mathilde Caron, Hugo Touvron, Ishan Misra, Herv{\'{e}} J{\'{e}}gou, Julien
  Mairal, Piotr Bojanowski, and Armand Joulin.
\newblock Emerging properties in self-supervised vision transformers.
\newblock {\em CoRR}, abs/2104.14294, 2021.

\bibitem{Gibson1966}
James~J. Gibson.
\newblock {\em The senses considered as perceptual systems}.
\newblock Houghton Mifflin, Boston, 1966.

\bibitem{plasticq}
James~J. Gibson.
\newblock {\em The Ecological Approach to Visual Perception}.
\newblock Houghton Mifflin, Boston, Boston 1979.

\bibitem{Binet1916}
A.~Binet and T.~Simon.
\newblock {\em The Development of Intelligence in Children: The Binet?Simon
  Scale}.
\newblock Williams \& Wilkins, 1916.

\bibitem{meloni2021sailenv}
Enrico Meloni, Luca Pasqualini, Matteo Tiezzi, Marco Gori, and Stefano Melacci.
\newblock Sailenv: Learning in virtual visual environments made simple.
\newblock In {\em 2020 25th International Conference on Pattern Recognition
  (ICPR)}, pages 8906--8913. IEEE, 2021.

\bibitem{DBLP:journals/corr/abs-2007-04954}
Chuang Gan, Jeremy Schwartz, Seth Alter, Martin Schrimpf, James Traer,
  Julian~De Freitas, Jonas Kubilius, Abhishek Bhandwaldar, Nick Haber, Megumi
  Sano, Kuno Kim, Elias Wang, Damian Mrowca, Michael Lingelbach, Aidan Curtis,
  Kevin~T. Feigelis, Daniel~M. Bear, Dan Gutfreund, David~D. Cox, James~J.
  DiCarlo, Josh~H. McDermott, Joshua~B. Tenenbaum, and Daniel L.~K. Yamins.
\newblock Threedworld: {A} platform for interactive multi-modal physical
  simulation.
\newblock {\em CoRR}, abs/2007.04954, 2020.

\end{thebibliography}
\bibliographystyle{unsrt}}






\end{document}